\pdfoutput=1

\documentclass[11pt]{article}

\usepackage[final]{acl}

\usepackage{times}
\usepackage{latexsym}
\usepackage{amsmath} 
\usepackage{booktabs}
\usepackage{longtable}
\usepackage{multirow}
\usepackage{array}

\usepackage[T1]{fontenc}

\usepackage[utf8]{inputenc}

\usepackage{microtype}

\usepackage{inconsolata}

\usepackage{graphicx}

\usepackage{tcolorbox}

\newcommand\blfootnote[1]{%
  \begingroup
  \renewcommand\thefootnote{}\footnote{#1}%
  \addtocounter{footnote}{-1}%
  \endgroup
}

\title{Rationale-Guided Retrieval Augmented Generation\\for Medical Question Answering}

\author{
    Jiwoong Sohn$^1$ \quad Yein Park$^1$ \quad Chanwoong Yoon$^1$ \quad Sihyeon Park$^1$ \\
    \textbf{Hyeon Hwang$^1$} \quad \textbf{Mujeen Sung$^2$} \quad 
    \textbf{Hyunjae Kim$^{1,\dagger}$} \quad \textbf{Jaewoo Kang$^{1,3,\dagger}$} \\
    $^1$Korea University \quad $^2$Kyung Hee University \quad $^3$AIGEN Sciences \\
    \texttt{\{jw\_sohn, hyunjae-kim, kangj\}@korea.ac.kr} \\
}

\begin{document}
\maketitle
\begin{abstract}
Large language models (LLM) hold significant potential for applications in biomedicine, but they struggle with hallucinations and outdated knowledge.
While retrieval-augmented generation (RAG) is generally employed to address these issues, it also has its own set of challenges: (1) LLMs are vulnerable to irrelevant or unhelpful context, (2) medical queries are often not well-targeted for helpful information, and (3) retrievers are prone to bias toward the specific source corpus they were trained on. 
In this study, we present RAG$^2$ (RAtionale-Guided RAG), a new framework for enhancing the reliability of RAG in biomedical contexts. 
RAG$^2$ incorporates three key innovations: a small filtering model trained on perplexity-based labels of rationales, which selectively augments informative snippets of documents while filtering out distractors; LLM-generated rationales as queries to improve the utility of retrieved snippets; a structure designed to retrieve snippets evenly from a comprehensive set of four biomedical corpora, effectively mitigating retriever bias. 
Our experiments demonstrate that RAG$^2$ improves the state-of-the-art LLMs of varying sizes, with improvements of up to 6.1\%, and it outperforms the previous best medical RAG model by up to 5.6\% across three medical question-answering benchmarks. Our code is available at \href{https://github.com/dmis-lab/RAG2}{https://github.com/dmis-lab/RAG2}

\end{abstract}

\blfootnote{\textsuperscript{$\dagger$}Corresponding authors.}

\section{Introduction}

Large language models (LLM)~\cite{OpenAI_2023,saab2024capabilities, llama3modelcard} have demonstrated remarkable performance across various tasks in biomedicine, including USMLE-style question-answering (QA) benchmarks~\cite{jin2021disease}.\footnote{The USMLE or United States Medical Licensing Examination is a standardized test that is required for obtaining a medical license in the United States.} Despite their state-of-the-art performance, LLMs face challenges limiting their adoption in high-stakes areas. A key concern is hallucination, where the model produces information that sounds plausible but is incorrect~\cite{maynez2020faithfulness,huang2023survey}. 
Additionally, updating models' knowledge is resource-intensive, making it difficult to maintain current medical information for application~\cite{zhang2023large,kasai2024realtime}.
Retrieval-augmented generation (RAG)~\cite{lewis2020retrieval} has emerged as a promising solution to address these limitations by reducing hallucinations and ensuring models provide up-to-date information through the integration of trustworthy documents into their input.

\begin{figure*}[t]
\centering
\includegraphics[width=\textwidth,trim=1 1 1 1]{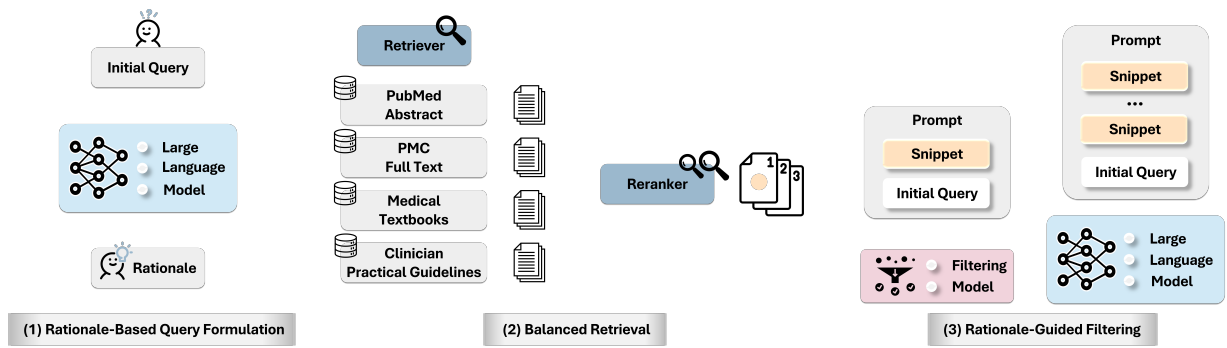}
\caption{
Our RAG$^2$ framework comprises the following three steps.
(1) {Rationale-based query formulation}: the model-generated rationale is used as the query for evidence retrieval instead of the initial query.
(2) {Balanced retrieval}: evidence snippets are retrieved in equal amounts from four corpora: two large corpora, PubMed and PMC, and two smaller but crucial corpora, textbooks and clinical guidelines. Subsequently, a reranker is used to rerank the retrieved snippets by cross-encoding the initial query and each snippet.
(3) {Rationale-guided filtering}: a  filtering model determines whether incorporating the retrieved snippets into the LLM prompt can help the model answer correctly and with higher confidence (as indicated by reduced perplexity).}
\label{figure:model_overview}
\end{figure*}

However, applying RAG in the biomedical domain presents unique challenges. First, irrelevant or incorrect context can mislead LLMs.~\cite{wu2024clasheval}.
In biomedical texts, the complexity and specialized terminology heighten the risk of retrieving documents that are not relevant or unhelpful.
Second, medical queries often struggle to target useful information. Some questions that include extensive patient details, such as medical histories and symptoms, may overwhelm retrieval systems, making it difficult to identify critical diagnostic clues. Conversely, overly brief questions often lack necessary context and depend on implicit medical knowledge, further complicating effective retrieval.
Lastly, retriever bias~\cite{chen2021evaluating,dai2024neural} also presents a significant challenge, as larger medical corpora often overshadow smaller, specialized ones that may contain critical, up-to-date information. 
For instance, MedCPT~\cite{jin2023medcpt} may show a preference for PubMed documents over clinical guidelines and medical textbooks, a tendency likely stemming from its training on the PubMed corpus.

Building upon these insights, we propose RAG$^2$ (RAtionale-Guided Retrieval Augmented Generation), a novel framework that improves the reliability of RAG in biomedical applications (see Figure~\ref{figure:model_overview}). 
First, we introduce a rationale-guided filtering method, training a Flan-T5 model~\cite{chung2024scaling} using labels derived from the differences in perplexity between rationales with and without retrieved documents. 
This model estimates the snippets' informativeness for the base LLM, selectively augmenting useful information while filtering out potential distractors. 
Next, our method substitutes medical questions with LLM-generated rationales as queries using chain-of-thought prompting~\cite{wei2022chain,jiang2023active,kang2024knowledge}. 
These rationale-based queries help identify key components through systematic problem-solving and also enable query expansion for brief questions~\cite{jagerman2023query,wang2023query2doc}.
Finally, we implement a balanced retrieval strategy that retrieves snippets equally from diverse biomedical corpora. 
This approach mitigates the potential bias present in dense retrievers by promoting a more comprehensive retrieval process across multiple information sources, regardless of their size or level of training exposure.

We evaluate our RAG$^2$ framework using three closed-book medical QA benchmarks—MedQA~\cite{jin2021disease}, MedMCQA~\cite{pal2022medmcqa}, and MMLU-Med~\cite{hendrycksmeasuring}—where no oracle documents are present.
Our method significantly enhances the average accuracy of state-of-the-art LLMs of varying sizes, including Llama-3-8B-Instruct~\cite{llama3modelcard}, Meerkat-7B~\cite{kim2024small}, and GPT-4o~\cite{openai2024gpt4o}.
Additionally, it consistently improves the performance of baseline LLMs by up to 6.1\% across three medical QA datasets and outperforms four recently developed RAG baselines, achieving a notable 5.6\% improvement over MedRAG~\cite{xiong-etal-2024-benchmarking} with the Llama-3-8B-Instruct model~\cite{llama3modelcard}.

Our contributions are as follows:
\begin{enumerate}
    \item We introduce RAG$^2$, a novel retrieval-augmented generation framework for medical QA. Our method incorporates advanced methods in query formulation, retrieval, and filtering processes that address core limitations in traditional RAG.
    \item Our rationale-guided filtering method employs a small language model trained with data labeled from differences in rationale perplexity to assess document utility. This approach not only enhances efficiency but also introduces a novel aspect to our methodology.
    \item Across three medical QA benchmarks, we demonstrate that the proposed RAG$^2$ framework significantly outperforms previous methods, including large commercial, small open-source, and medical-specialized models, with an average accuracy improvement of 6.1\%, 3.8\%, and 0.9\%, respectively.
    \end{enumerate}

\section{Related Work}

\subsection{Retrieval-Augmented Generation}
Basic RAG approaches~\cite{lewis2020retrieval, gao2023retrieval} involve the straightforward addition of retrieved documents to input queries.
Advanced methods train LLMs to perform RAG adaptively~\cite{wang2023learning,zhang2024raft, gan2024similarity}. 
Self-RAG~\cite{asai2023self}, for instance, trains an LLM with synthetic data from GPT-4 to determine whether to use the retrieved document or not.
However, these methods generally require substantial computational resources and API costs, making model updates challenging.

Other approaches focus on training smaller models separately from the base LLM. 
For instance, ARES \cite{saad2024ares} employs three small language models trained on a mix of synthetic and human-labeled data to assess context relevance, answer faithfulness, and answer relevance. 
CRAG~\cite{yan2024corrective} uses an iterative approach with a small evaluator model but it still relies on GPT-3.5 for query rewriting. 
Adaptive-RAG \cite{jeong2024adaptive} assesses query complexity based on changes in the correctness of an LLM's response, but can only utilize to queries that are answered correctly, overlooking the nuanced effects of retrieved documents beyond basic accuracy. 
In contrast, our method measures perplexity differences in the rationales generated by the base LLM, providing the filtering model with more detailed training signals.

Several RAG methods leverage the rationales of models to enhance RAG. 
Speculative RAG~\cite{wang2024speculative} generates multiple pseudo-answers with rationales and uses a large verifier model to select the best answer. 
RAT~\cite{wang2024rat} iteratively performs retrieval and refines the initial rationale to obtain the final answer. 
However, these approaches come with significant downsides, such as high computational costs due to the use of large models for training and multiple LLM calls, as well as increased latency from the iterative retrieval and decision-making processes. 

Uncertainty-based approaches like FLARE ~\cite{jiang2023active} and SEAKR~\cite{yao2024seakr} dynamically guide RAG by using confidence levels to make retrieval decisions.
However, they both rely on numerous iterative retrieval and pseudo-generations, leading to significant computational costs as well. 
Our approach also incorporates uncertainty but restricts its application to the training stage, thus avoiding the expensive iterative processes during the inference stage.

\subsection{Medical RAG}
Researchers have proposed several methods to enhance biomedical information retrieval and RAG frameworks. 
MedCPT~\cite{jin2023medcpt} introduced off-the-shelf retriever and reranker models tailored for the biomedical domain. 
It overcame the limited availability of query-article annotations in medicine domain by using 255 million user click logs from PubMed.
MedRAG~\cite{xiong-etal-2024-benchmarking} is an RAG toolkit to comprehensively integrate indexing, retrieval, and reranking processes. It extracts documents from the MedCorps corpus using a hybrid approach that integrates both sparse and dense retrievers.
Self-BioRAG~\cite{jeong2024improving} adopts the concept of Self-RAG to the domain to handle complex medical queries more effectively.

Concurrent with our work, Bailicai~\cite{long2024bailicai} introduced features, including self-knowledge boundary identification and directed acyclic graph-based task decomposition. 
Additionally, i-MedRAG~\cite{xiong2024improving} incorporates iterative follow-up query generation. We would like to note that, with the exception of MedRAG, all other approaches necessitate fine-tuning LLMs or multiple iterative RAG executions.
In contrast, our method highlights improved training efficiency and run-time performance by training only a small filtering model and relying on single-step retrieval.

\section{Method}

We start by explaining the training objective of RAG models. 
Next, we provide a detailed breakdown of the pipeline, focusing on three key steps: (1) rationale-guided filtering, (2) rationale-based query formulation, and (3) balanced retrieval. Figure~\ref{figure:model_overview} presents an overview of the entire model.

\subsection{Objective}

Let $\mathbf{x} = [x_1, \dots, x_N]$ represent a natural language query consisting of $N$ tokens, and let $\mathbf{y} = [y_1, \dots, y_M]$ represent the output sequence consisting of $M$ tokens.
An LLM is trained to minimize the probability of generating the output sequence $\mathbf{y}$ by maximizing the likelihood of the following conditional probabilities for each token $y_t$ as follows:
\begin{equation}
    p(\mathbf{y}|\mathbf{x}, \mathbf{D}^*) = \prod_{t=1}^{M} p(y_t \mid \mathbf{y}_{<t}, \mathbf{x}, \mathbf{D}^*),
\end{equation}
where \( \mathbf{y}_{<t} \) represents the sequence of tokens generated up to time step $t-1$, and $\mathbf{D}^* = [\mathbf{d}_1, \dots, \mathbf{d}_{k}]$ represents the top-k retrieved documents that are most relevant to the query.
The retrieval process involves embedding both the query and documents in a shared latent space using a query encoder \( f_{\text{q}} \) and a document encoder \( f_{\text{d}} \).
The most relevant documents are then selected based on their inner product as follows:
\begin{equation}
    \mathbf{D}^* = \operatorname{Top-k} \left(\langle f_{\text{q}}(\mathbf{x}), f_{\text{d}}(\mathbf{d}) \rangle | \mathbf{d} \in \mathcal{C} \right),
\label{equation:topk}
\end{equation}
where $\mathcal{C}$ represents the set of candidate documents.

\begin{figure}[t]
\centering
\includegraphics[width=0.95\columnwidth]{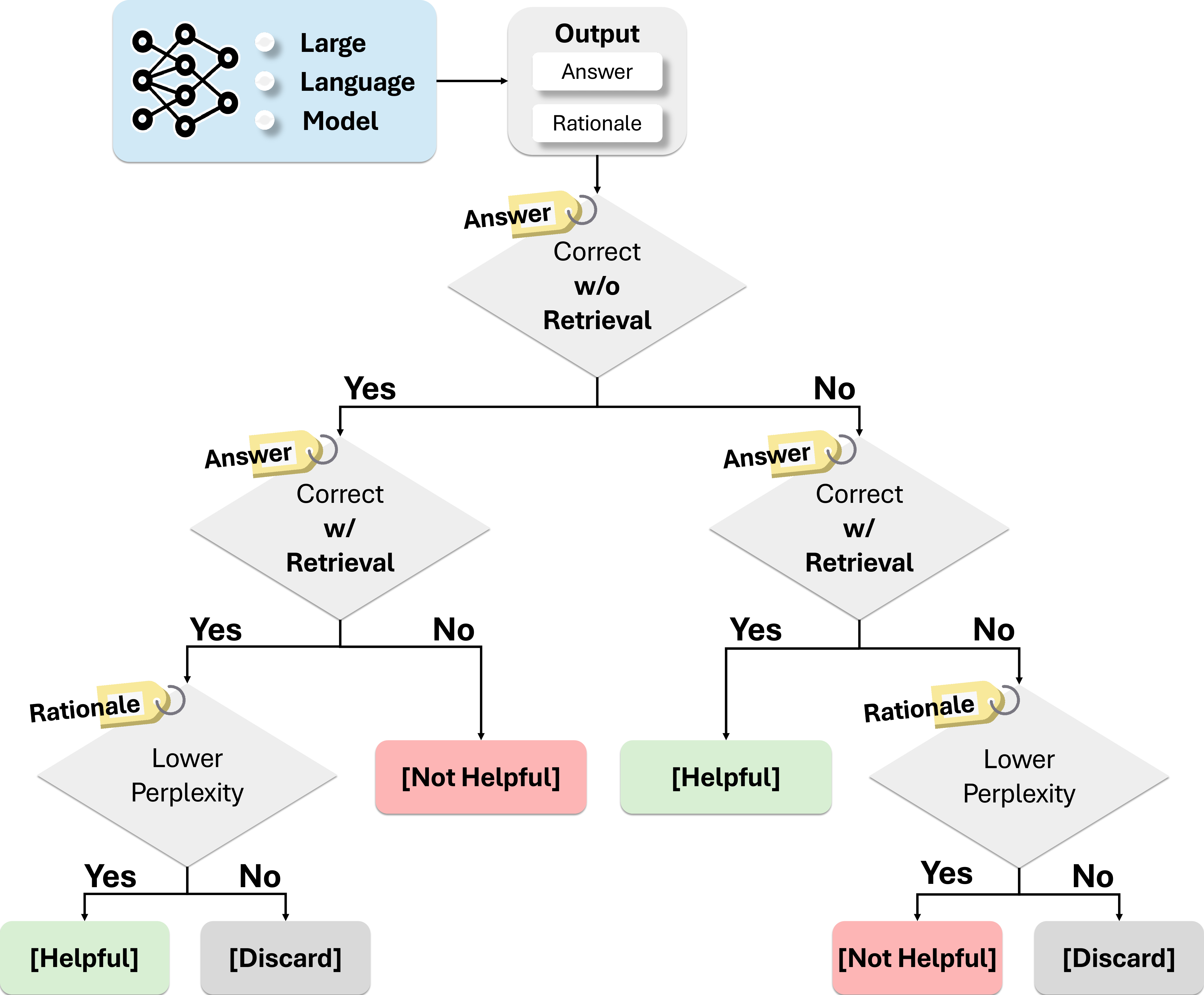}
\caption{
Illustration of the data annotation process used to train our rationale-guided filtering model.
}
\label{figure:data_labeling}
\end{figure}

\subsection{Rationale-Guided Filtering}

Retrieval mechanisms do not always pick the most helpful documents as similarity alone does not ensure a positive contribution to the model's output. 
To filter out distractors and obtain only helpful document, we train a filtering model to evaluate the retrieved documents before they are presented to the LLM. 
Our method begins by analyzing the LLM’s outputs when answering questions under two conditions: (1) using its intrinsic parametric knowledge, and (2) using external information retrieved from a knowledge base. 
If the model successfully answers a question with the aid of retrieved documents but fails to do so independently, the documents are labeled as ``helpful''; otherwise, they are labeled as ``not helpful.''
This labeling strategy was similarly adopted in previous works~\cite{jeong2024adaptive}, but it fails to fully capture the document utility, particularly in cases where the model’s accuracy remains unchanged regardless of whether the document is included or not.

To address this, we calculate the difference in perplexity, \(\Delta \text{PPL}\), to determine the document's impact on the model's confidence as follows:
\begin{equation}
    \Delta \text{PPL} = \text{PPL}(\mathbf{x}) - \text{PPL}(\mathbf{x}, \mathbf{d}) \geq \tau,
\end{equation}
where $\mathbf{d}$ is the target document and $\tau$ represents a threshold set to select the top percentage of perplexity differentials. 
The perplexity for the input query $\mathbf{x}$ and the document $\mathbf{d}$ are calculated as follows:
\begin{equation}
    \begin{aligned}
    &\text{PPL}(\mathbf{x}) = \exp\left(-\frac{1}{L} \sum_{i=0}^{L-1} \log P(x_i \mid \mathbf{x}_{<i})\right),\\
    &\text{PPL}(\mathbf{x}, \mathbf{d}) = \exp\left(-\frac{1}{L} \sum_{i=0}^{L-1} \log P(x_i \mid \mathbf{x}_{<i}, \mathbf{d})\right).
    \end{aligned}
\end{equation}

It is important to note that while lower perplexity (i.e., higher model confidence) generally correlates with higher accuracy, this relationship is not always consistent. 
Perplexity alone cannot reliably guarantee accuracy, making it necessary to introduce a threshold for more precise labeling. 
Through our validation process, we determined that setting the threshold value $\tau$ to the top 25\% of perplexity differentials consistently yielded the best performance and was therefore fixed across all our experiments.

Figure~\ref{figure:data_labeling} illustrates the data annotation process. By applying this systematic labeling approach, we generate a perplexity-stratified dataset derived from the base model’s outputs, which is subsequently used to train a Flan-T5-large model as our filtering model. This method not only ensures that the filtering model effectively assesses the utility of retrieved documents for the base model, allowing only relevant and beneficial information to be integrated into the LLM, but also helps address the challenge of scarce annotated data in the medical domain by utilizing model-generated labels instead of relying solely on human annotations.

\subsection{Rationale-Based Query Formulation}
\label{subsec:query_formulation}

As we discussed in the introduction, the initial query, $\mathbf{x}$, can be verbose or too brief, increasing the complexity of medical QA.
To address this, we use model-generated rationales as queries to search for relevant information~\cite{wang2023query2doc,kang2024knowledge}.
Following~\citet{kim2024small}, we extract the LLM's rationales using the following chain-of-thought prompt: 
\begin{tcolorbox}[colback=gray!3,colframe=black]
The following are multiple choice questions about medical knowledge. Solve them in a step-by-step fashion, starting by summarizing the available information. Output your explanation and single option from the given options as the final answer.\\
Here is the question: \texttt{[initial\_query]}
\end{tcolorbox}

Here, we replace \texttt{[initial\_query]} with the initial query $\mathbf{x}$ and obtain the model's response.
We search for document snippets solely using the rationale, excluding the initial query. 
Including both the initial query and the rationale exceeds the maximum length of the retriever and, as confirmed by our initial experiments, leads to suboptimal performance.
Also, we note that the same LLM is used both for rationale generation and QA. 
In our experiments, we show that rationale queries can still be effective even with smaller open-source models (see Table~\ref{table:main_results} in Experiments). 

\subsection{Balanced Retrieval}
Searching documents from a broad corpus can provide high coverage, but it can lead the retriever to favor dominant corpora and popular information~\cite{chen2021evaluating,kim2023automatic}.
As a result, smaller yet essential corpora may be underrepresented.
To address this issue, we use balanced retrieval, a simple yet effective method. 
This approach extracts an equal number of documents from each corpus, ensuring that all corpora are represented more evenly compared to existing methods~\cite{xiong-etal-2024-benchmarking}.
After the balanced retrieval stage, we further refine the selection of relevant documents. 
We use MedCPT~\cite{jin2023medcpt}, an off-the-shelf reranker, which encodes the original query along with each document to determine a relevance score. 
This process prioritizes documents that are more closely aligned with the query by reranking the results initially retrieved from multiple sources in a balanced manner, enhancing overall performance.

\section{Experiments}

\subsection{Datasets}
We utilize three widely recognized medical QA datasets. (1) MedQA~\cite{jin2021disease} consists of USMLE-style questions curated by medical examination experts from various medical question banks. (2) MedMCQA~\cite{pal2022medmcqa} comprises exam questions from the two Indian affiliations, AIIMS (All India Institute of Medical Sciences) and NEET PG (National Eligibility cum Entrance Test for Post Graduate courses).
(3) MMLU-Med, a subset of the MMLU dataset~\cite{hendrycksmeasuring}, consists of questions for six biomedical subjects—clinical knowledge, medical genetics, anatomy, professional medicine, college biology, and college medicine—spanning from high school to professional-level knowledge.
All of these datasets consist of multiple-choice questions with four answer options. Detailed statistics about the datasets can be found in Table~\ref{table:dataset}.

\begin{table}[t]
\centering
\footnotesize
\begin{tabular}{lrrr}
\toprule
\textbf{Dataset} & \textbf{Training} & \textbf{Validation} & \textbf{Test} \\
\midrule
MedQA & 10,178 & 1,272 & 1,273 \\
MedMCQA & 182,822 & 4,183 & 6,150 \\
MMLU-Med & - & - & 1,089 \\
\midrule
\end{tabular}
\caption{The number of question-answer pairs in the three medical QA datasets, MedQA~\cite{jin2021disease}, MedMCQA~\cite{pal2022medmcqa}, and MMLU-Med~\cite{hendrycksmeasuring}.}
\label{table:dataset}
\end{table}

\begin{table*}[t]
\centering
\footnotesize
\begin{tabular}{lcccc}
\toprule
\textbf{Model} & \textbf{MedQA} & \textbf{MedMCQA} & \textbf{MMLU-Med} & \textbf{Average} \\
\midrule
\multicolumn{5}{l}{\textbf{\textit{Open-source LLMs}} (zero- or few-shot)} \\
\midrule
Llama-2-7B~\cite{touvron2023llama} (3-shot) & 35.2 & 36.3 & 46.3 & 39.3 \\
Mistral-7B-Instruct~\cite{jiang2023mistral} (0-shot) & 41.1 & 40.2 & 55.8 & 45.7 \\
\midrule
\underline{Llama-3-8B-Instruct}~\cite{llama3modelcard} (0-shot) & 57.7 & 53.5 & 69.5 & 60.2 \\
+ MedCPT~\cite{jin2023medcpt} ($k=1$) &55.3&51.3&65.8&57.5 \\
+ MedCPT+Rationale ($k=1$) & 58.0 &52.1&70.3&60.1 \\
+ MedRAG~\cite{xiong-etal-2024-benchmarking} & 56.4 & 56.6 & 69.2&60.7   \\
+ query2doc~\cite{wang2023query2doc} ($k=1$) & 54.3 & 50.0 & 58.5 & 54.3   \\
+ Adaptive-RAG \cite{jeong2024adaptive}  & 57.3 & 53.1 & 70.3 & 60.2 \\
+ InstructRAG-ICL (2-shot)~\cite{wei2024instructrag} & 55.5 & 55.7 &71.9&61.8 \\
+ RAG$^2$ (\textbf{Ours}) & \textbf{64.6}  & \textbf{59.4}  & \textbf{74.8}  &\textbf{66.3}  \\
\midrule
\multicolumn{5}{l}{\textbf{\textit{Medical LLMs}} (fine-tuned)} \\
\midrule
MediTron-7B~\cite{chen2023meditron} & 50.2 & 57.9 & 56.7 & 54.9 \\
BioMistral-7B~\cite{labrak2024biomistral} & 54.3 & 61.1 & 64.6 & 60.0 \\
\midrule
\underline{Meerkat-7B}~\cite{kim2024small} & 71.2 & 60.8 & 73.8 & 68.6 \\
+ MedCPT~\cite{jin2023medcpt} ($k=1$) & 71.8& 57.9&74.0&67.9 \\
+ MedCPT+Rationale ($k=1$) & 73.3&58.4&75.7&69.1\\
+ MedRAG~\cite{xiong-etal-2024-benchmarking} & 67.9 & 60.6 &76.1  & 68.2  \\
+ query2doc~\cite{wang2023query2doc} ($k=1$) & 70.3 & 53.8 & 73.6 & 65.9  \\
+ Adaptive-RAG \cite{jeong2024adaptive} & 71.4 & 60.5 & 74.0 & 68.6 \\
+ InstructRAG-ICL (2-shot)~\cite{wei2024instructrag} &65.8&53.2&63.7&60.9 \\
+ RAG$^2$ (\textbf{Ours}) & \textbf{75.6} & \textbf{63.0} &\textbf{78.7}   & \textbf{72.4}  \\
\midrule
\multicolumn{5}{l}{\textbf{\textit{Commercial LLMs}} (zero- or few-shot)} \\
\midrule
GPT-3.5~\cite{openai2022chatgpt} (5-shot)& 53.6 & 51.0 & 67.3 & 57.3 \\
GPT-4~\cite{OpenAI_2023} (5-shot) & 81.4 & 72.4 & 87.1 & 80.3 \\
\midrule
\underline{GPT-4o} (0-shot) & 88.5 & 76.7 & \textbf{92.8} & 86.0 \\
+ MedCPT~\cite{jin2023medcpt} ($k=1$) &86.6&72.5&90.1&83.1 \\
+ MedCPT+Rationale ($k=1$) &87.3&74.7&90.2&84.1 \\
+ MedRAG~\cite{xiong-etal-2024-benchmarking} & 88.3 & 75.9 &92.4  & 85.5 \\
+ query2doc~\cite{wang2023query2doc} ($k=1$) & 89.1 & 73.9 & 91.5 & 84.8   \\
+ Adaptive-RAG \cite{jeong2024adaptive} & 88.5 & 76.7 & 92.5 & 85.9 \\
+ InstructRAG-ICL (2-shot)~\cite{wei2024instructrag} &87.7&73.5&90.0&85.6 \\
+ RAG$^2$ (\textbf{Ours}) & \textbf{91.1} & \textbf{77.2} &92.5  &\textbf{86.9}   \\
\bottomrule
\end{tabular}
\caption{
Performance (accuracy) of LLMs and RAG models on the three medical QA benchmarks. 
RAG methods are applied to the underlying models, specifically Llama-3-8B-Instruct, Meerkat-7B, and GPT-4o.
The best scores for each benchmark are highlighted in bold. 
}
\label{table:main_results}
\end{table*}

\begin{figure*}[t]
\centering
\includegraphics[width=1\textwidth]{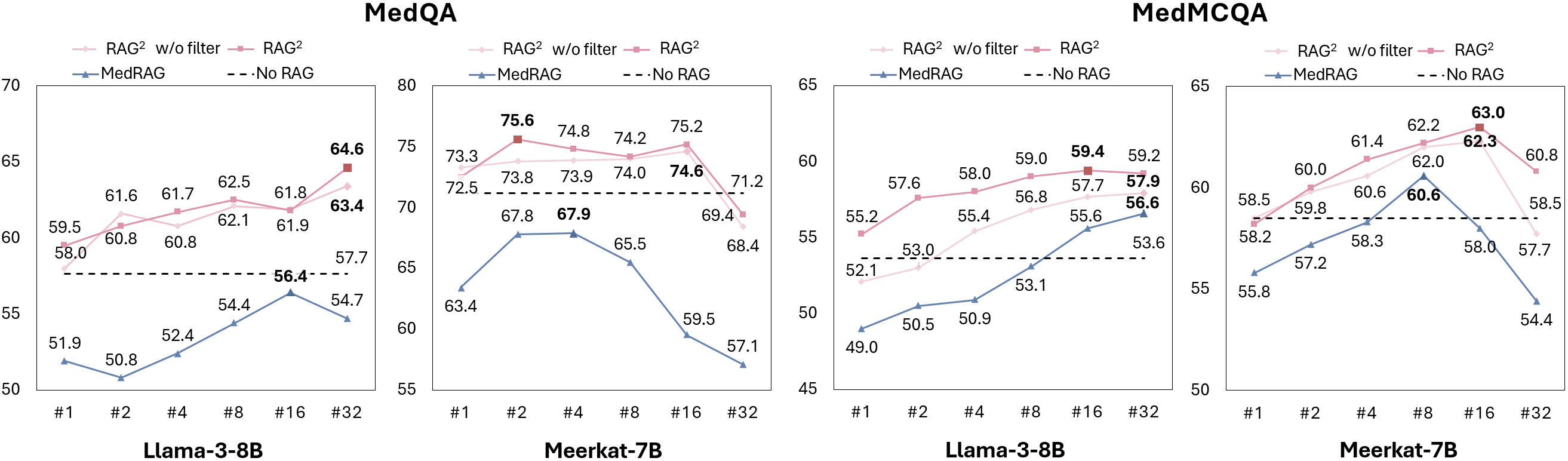} 
\caption{
The accuracy (y-axis) of Llama-3-8B-Instruct and Meerkat-7B with/without different RAG models when varying the number of top-k snippets (x-axis). 
``RAG$^2$ w/o filter'' is our RAG framework only that does not the filtering model.
The performance on the MMLU-Med dataset can be found in the appendix.
}
\label{figure:topk}
\end{figure*}

\subsection{Models}
\paragraph{Large Language Models}
We use state-of-the-art baseline models from three categories: (1) open-source LLMs (Llama-2-7B~\cite{touvron2023llama}, Mistral-7B-Instruct~\cite{jiang2023mistral}, Llama-3-8B-Instruct~\cite{llama3modelcard}), (2) medical LLMs (MediTron-7B~\cite{chen2023meditron}, BioMistral-7B~\cite{labrak2024biomistral}, Meerkat-7B~\cite{kim2024small}), and (3) commercial LLMs (GPT-3.5~\cite{openai2022chatgpt}, GPT-4~\cite{OpenAI_2023}, GPT-4o).
From these categories, we select the top-performing models based on benchmark performance as our backbone models.
(1) Llama-3-8B-instruct\footnote{\url{https://huggingface.co/meta-llama/Meta-Llama-3-8B-Instruct}}, part of the Llama series~\cite{touvron2023llama}, is the state-of-the-art open-source model. We use this model as-is, without further fine-tuning on the target datasets.
For simplicity, we refer to it as Llama-3-8B.
(2) Meerkat-7B is initialized using the Mistral-7B weights~\cite{jiang2023mistral} and further instruction-tuned with rationales generated by GPT-4. The model is then fine-tuned specifically on the MedQA and MedMCQA datasets.
(3) GPT-4o is the latest version of OpenAI's closed-source LLM, demonstrating state-of-the-art performance across various tasks.
We apply our RAG method to these models to assess whether our RAG method can further enhance their high performances without the need for fine-tuning.

\paragraph{RAG Methods}
We conduct a comprehensive comparison between the state-of-the-art RAG models and our proposed method:
(1) MedCPT~\cite{jin2023medcpt} is a retriever pre-trained using user search logs from PubMed.
(2) MedCPT+Rationale refers to the MedCPT retriever enhanced with model-generated rationale queries.
(3) MedRAG~\cite{xiong-etal-2024-benchmarking} employs four different sparse and dense retrievers to gather information from a collection of corpora called MedCorp, then reranks the retrieved results using Reciprocal Rank Fusion (RRF)~\cite{cormack2009reciprocal}.
(4) query2doc~\cite{wang2023query2doc} instructs LLMs to make pseudo-documents for query expansion.
(5) Adaptive-RAG~\cite{jeong2024adaptive} trains a filtering model with labels obtained from correctly answered queries with and without RAG.
(6) InstructRAG~\cite{wei2024instructrag} employs explanatory rationales generated using the training set as few-shot demonstrations for in-context learning. 
We do not include Self-BioRAG~\cite{jeong2024improving} in our experiments for two reasons. First, our goal is to present an efficient RAG approach that trains only a small filtering model, without the need to train the LLM itself.
While Self-BioRAG trains the Llama-2 7B and 13B models using data generated by GPT-4, we opt to train a filtering model based on Flan-T5-large, which has only 770 million parameters~\cite{chung2024scaling}. This smaller model can be trained on a single RTX 3090 24G GPU.
Second, we aim for computational efficiency through single-pass generation, in contrast to Self-BioRAG, which generates multiple candidate answers and selects the highest-scoring response. 

\paragraph{Implementation Details}
In the experiment, the MedCPT and MedCPT+Rationale models rely solely on the top-1 snippet to investigate performance variations across different query types. 
The query2doc model has not been assessed in a closed-book QA setting. For consistency with MedCPT+Rationale, we set $k=1$. 
For Adaptive-RAG and InstructRAG, we apply the specified hyperparameters of $k=1$ and $k=5$, respectively. 
MedRAG and RAG$^2$ use the optimal top-k values, determined through validation.
We apply the same index and retriever to the MedCPT, query2doc, Adaptive-RAG, and InstructRAG as those used for RAG$^2$, whereas MedRAG uses its own methods.
We train our filtering and Adpative-RAG models using QA pairs from MedQA and MedMCQA.
For Adaptive-RAG, we label training set queries as ``simple'' if the model could answer them correctly without retrieval and ``complex'' if it could only answer them correctly with RAG. 
We then train a small binary classifier using these labeled queries as input.
Since MMLU-Med does not provide training data, we use the MedMCQA training data to train the filtering model. Details of the retrieval corpus are described in the appendix.

\subsection{Main Results}

Table~\ref{table:main_results} presents the accuracy of various LLMs, with various RAG baselines and RAG$^2$, across three medical QA benchmarks. The results indicate that our RAG$^2$ method consistently enhances the baseline LLMs, yielding an average score improvement of 6.1\%, 3.8\%, and 0.9\% for Llama-3-8B-Instruct, Meerkat-7B, and GPT-4o respectively. 
Additionally, the performance improvement on MMLU-Med is 5.3\% for Llama-3-8B and 4.9\% for Meerkat-7B, demonstrating the transferability of our filtering model to out-of-domain datasets.

In contrast, some baseline RAG methods show inconsistent results, with some even showing lower average scores.
This suggests that without advanced retrieval and filtering modules, RAG frameworks do not always guarantee improved performance, especially in the medical domain.
The performance of the query2doc model declines significantly with smaller models, likely due to their limited ability to generate effective pseudo-documents for answering queries. On the other hand, MedCPT+Rationale maintains strong performance across models of various capacities, suggesting that our rationale-based query approach is more robust and adaptable.

Additional experiments using an ensemble of different filtering models with GPT-4o on MedQA demonstrates a score increase to 91.6. However, as our study focuses on single-pass generation, we do not include ensemble results in Table~\ref{table:main_results}.

Although Adaptive-RAG shares similarities with our RAG$^2$ model in that using a filtering model with the training data, the results differ significantly.
This is because the perplexity-based labels we use offer a more informative and fine-grained training signal to the filtering model compared to the coarse correct-or-incorrect labels used by Adaptive-RAG.
We discuss this further in Section~\ref{subsec:ablation}.

\section{Analysis}
\subsection{Ablation Study}
\label{subsec:ablation}
\paragraph{The Number of Retrieved Snippets}
Figure~\ref{figure:topk} illustrates the performance of various RAG methods across different top-k values, using Llama-3-8B-Instruct and Meerkat-7B as backbone LLMs. 
MedRAG demonstrates inconsistent performance, often underperforming compared to LLMs without retrieval capabilities. 
In contrast, the use of rationale queries generally enhances the performance of the baseline LLMs, leading to improvements of up to 5.7\% on MedQA and 4.3\% on MedMCQA for Llama-3-8B-Instruct, and up to 3.4\% on MedQA and 3.8\% on MedMCQA for Meerkat-7B. 
Further application of rationale-guided filtering results in additional performance gains, with Llama-3-8B achieving up to 6.9\% on MedQA and 5.8\% on MedMCQA, and Meerkat-7B reaching up to 4.4\% on MedQA and 4.5\% on MedMCQA.

For Llama-3-8B-Instruct, the performance tends to improve as the top-k value increases, while Meerkat-7B shows a peak in performance followed by a decline. 
This difference could be attributed to the models' trained context lengths: the Llama model is trained with a maximum context length of 8K, whereas Meerkat-7B is instruction-tuned with a 2K maximum length. This likely explains the performance drop when more snippets are provided. Nonetheless, across all top-k values, our RAG$^2$ method consistently outperforms MedRAG.

\begin{table}[t]
\centering
\footnotesize
\resizebox{\columnwidth}{!}{
\begin{tabular}{lcc}
\toprule
\textbf{Method} & \begin{tabular}[c]{@{}c@{}}\textbf{Med-}\\\textbf{QA}\end{tabular} & \begin{tabular}[c]{@{}c@{}}\textbf{Med-}\\\textbf{MCQA}\end{tabular} \\
\midrule
\underline{Llama-3-8B} (MedCPT w/o Filtering) & 55.3 & 51.3  \\
+ Adaptive-RAG~\cite{jeong2024adaptive} & 57.3 & 53.1   \\
+ InstructRAG-ICL~\cite{wei2024instructrag} & 55.5 & 55.7  \\
+ GPT-4o & 58.5 & \textbf{55.8} \\
+ Rationale-guided Filtering (\textbf{Ours}) & \textbf{58.6} & \textbf{55.8} \\
\bottomrule
\end{tabular}
}
\caption{
Performance of Llama-3-8B-Instruct depending on the filtering method applied.
Top-1 documents are used except for InstructRAG-ICL (top-5)
}
\label{table:filtering_methods}
\end{table}

\paragraph{Filtering Methods}
We evaluate how the performance of the baseline model, Llama-3-8B-Instruct, is affected by different filtering methods. To isolate the impact of filtering in this experiment, we fix the top-k value at 1 and used the original query instead of a rationale-based query. For InstructRAG, top-k value is 5 following the default setting. 
The ``GPT-4o'' filtering determines whether a retrieved document contributes to answering the question. 
Our filtering method uses the perplexity of the LLM’s rationale to annotate question-document pairs. 
As a result, this approach significantly improves Llama’s performance, surpassing Adaptive-RAG by 1.3\% on MedMQ and 2.7\% on MedMCQA, as shown in Table~\ref{table:filtering_methods}. Additionally, it achieves results similar to GPT-4o while avoiding its API costs.

\begin{table}[t]
\centering
\footnotesize
\begin{tabular}{lccc}
\toprule
\textbf{Method} & \textbf{MedQA} & \textbf{MedMCQA} & \textbf{MMLU-Med} \\
\midrule
\multicolumn{4}{l}{\underline{Llama-3-8B-Instruct}} \\
+ MedRAG & {51.9} & {49.0} & {65.1} \\
+ Ours & \textbf{55.3} & \textbf{51.3} & \textbf{65.8} \\
\midrule
\multicolumn{4}{l}{\underline{Meerkat-7B}} \\
+ MedRAG & {63.2} & {56.0} & {73.1} \\
+ Ours & \textbf{71.8} & \textbf{57.9} & \textbf{74.0} \\
\bottomrule
\end{tabular}
\caption{
Comparison with MedRAG~\cite{xiong-etal-2024-benchmarking} and our balanced retrieval.
Top-1 documents are used.
}
\label{table:comparison_with_medrag}
\end{table}

\paragraph{Effect of Balanced Retrieval}

Table~\ref{table:comparison_with_medrag} demonstrates that our balanced retrieval consistently outperforms MedRAG, highlighting the effectiveness of extracting information in a balanced manner from each corpus. We used the same initial query and top-1 document, without applying other methods, to solely compare the retrieval component. Additionally, further studies with corpus ablation, as well as ablation studies of balanced retrieval, are provided in the appendix to demonstrate how this approach mitigates retriever bias.

\begin{figure*}[t!]
\centering
\includegraphics[width=0.95\textwidth]{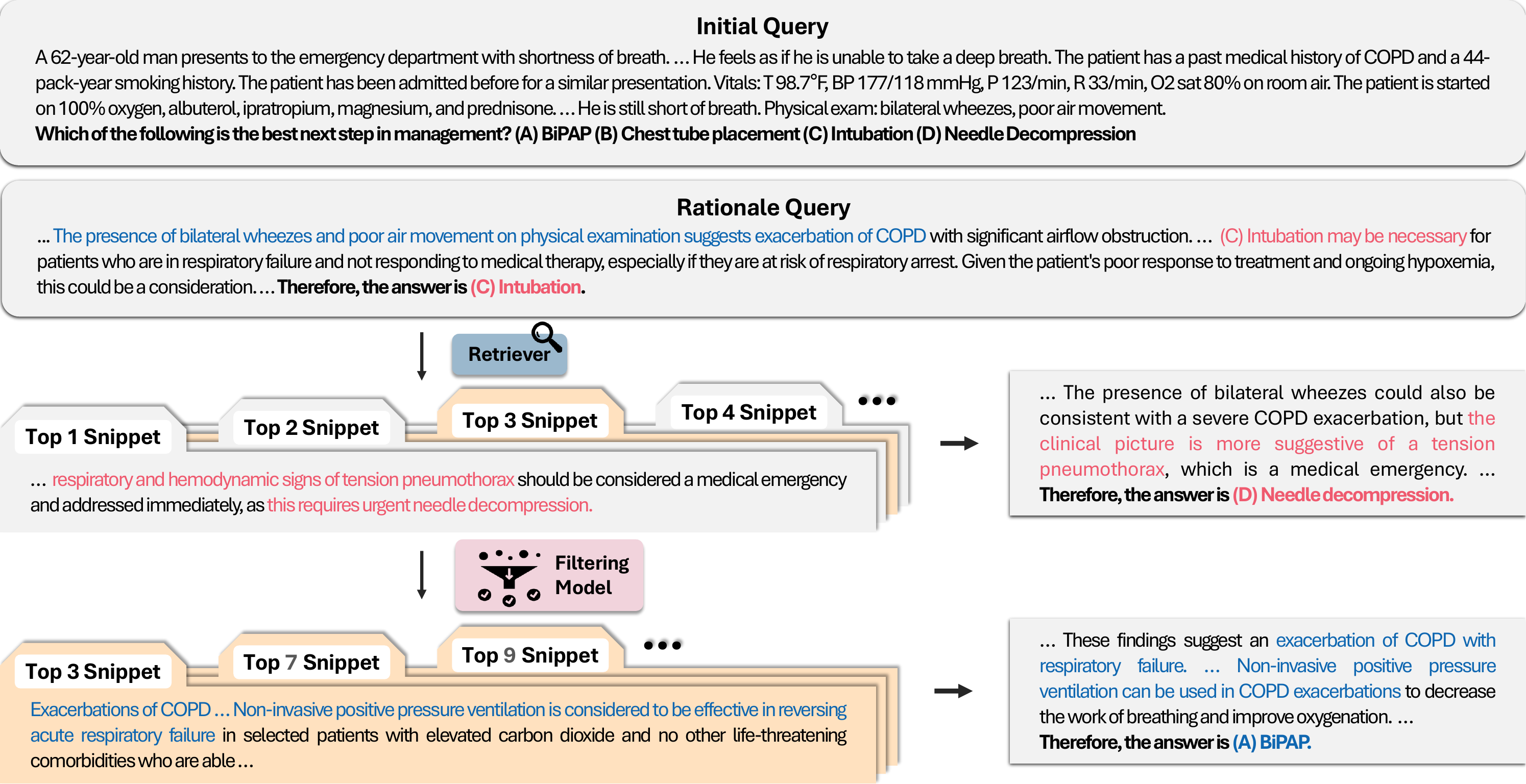} 
\caption{
This case study demonstrates the effectiveness of our rationale-guided filtering approach (see the main text for detailed descriptions).
Some content has been omitted due to space limitations.
}
\label{figure:case_study}
\end{figure*} 

\subsection{Quality of Rationale Queries} 
Table~\ref{table:rationale_query} highlights that rationale queries generated by higher-performing models consistently enhance the accuracy of LLMs on MedQA. Specifically, GPT-4o-generated rationales consistently yield the highest performance improvements across all models, followed by Meerkat-7B and Llama-3-8B.
High-quality rationales refine poorly targeted, complex medical questions into precise and informative queries, minimizing the retrieval of non-informative documents and improving overall retrieval effectiveness.

\begin{table}[t]
\centering
\resizebox{\columnwidth}{!}{
\begin{tabular}{lcc|c}
\toprule
\textbf{} & \multicolumn{3}{c}{\textbf{Rationale Query of}} \\
\cmidrule(lr){2-4}
\textbf{Model} & \multicolumn{1}{c}{\textbf{Llama-3-8B}} & \textbf{Meerkat-7B} & \textbf{GPT-4o} \\
\midrule
Llama-3-8B & 63.4 & 71.5 & \textbf{73.6} \\
Meerkat-7B & 71.3 & 74.6 & \textbf{78.8} \\
GPT-4o & 87.4 & 88.3 & \textbf{89.8} \\
\bottomrule
\end{tabular}
}
\caption{Performance comparison on MedQA of backbone models using rationale queries generated by Llama-3-8B-Instruct, Meerkat-7B, and GPT-4o.
}
\label{table:rationale_query}
\end{table}



\subsection{Case Study}
\label{subsec:casestudy}

Figure~\ref{figure:case_study} presents an example from the MedQA dataset involving a patient with a severe COPD exacerbation. 
Initially, the Meerkat-7B model recognizes the severity of the situation but fails to prioritize less invasive options like BiPAP (Option~A), which would be appropriate before considering intubation (Option~C). 
This outcome highlights the limitations of relying solely on the model’s parametric knowledge.
When applying RAG with the rationale query, the model is presented with a snippet containing information about tension pneumothorax and needle decompression. Although this information is medically accurate, it is irrelevant to the patient’s condition of COPD exacerbation. The presence of this extraneous data causes the model to incorrectly infer the possibility of a tension pneumothorax, leading to an inappropriate recommendation for needle decompression (Option~D).
After applying our filtering method to remove the irrelevant information about tension pneumothorax, the model correctly identifies the patient’s condition as severe COPD exacerbation. As a result, it selects BiPAP (Option A) as the appropriate next step.
This example underscores the critical role of document filtering in reliable RAG: the initial error stems from the model's distraction by irrelevant information, but proper filtering lead to the correct diagnosis and management plan. 
This case highlights how our method could improve the base LLM’s ability to deliver more accurate clinical guidance by prioritizing relevant snippets and minimizing the risk of misdiagnosis in real-world scenarios.

\section{Conclusion}
In this study, we presented RAG$^2$, a novel framework designed to enhance the reliability and performance of LLMs in biomedical QA tasks. By integrating rationale queries, balanced retrieval, and a rationale-guided filtering method, we created a comprehensive approach that effectively refined and optimized the entire RAG pipeline. Our experimental results consistently demonstrated that RAG$^2$ significantly improved the accuracy of various state-of-the-art LLMs across multiple medical QA benchmarks. This advancement marked an important step toward more reliable AI-assisted medical decision-making.

\section*{Limitation}
Our RAG$^2$ framework has several limitations. 
First, our model has only been tested in the biomedical domain, leaving its applicability to general domains unexamined. We plan to explore its performance in various general domains in future research.
Second, we used only one size of the Flan-T5 model as the filtering model; experimenting with different sizes or architectures may yield additional improvements.
Third, an incorrect rationale from the model might prompt the retriever to gather distracting evidence. Although this risk is limited, as errors generally make up only a small portion of the rationale and our filtering model can discard distractors, a thorough analysis and evaluation of this issue remains essential for future work.
Lastly, when creating perplexity-based labels, we evaluated each snippet individually, which may overlook the combined impact of multiple relevant snippets. Also, the Flan-T5 model can filter only one snippet at a time due to its limited context length. 
In future research, we plan to explore both a labeling approach that considers multiple related snippets and new encoding methods to overcome the structural limitations of the current filtering model.

\section*{Acknowledgements}
This research was supported by (1) the National Research Foundation of Korea (NRF2023R1A2C3004176, RS-2023-00262002), (2) the Ministry of Health \& Welfare, Republic of Korea (HR20C0021), and (3) ICT Creative Consilience Program through the Institute of Information \& Communications Technology Planning \& Evaluation(IITP) grant funded by the Korea government(MSIT)(IITP-2025-20200-01819). Mujeen Sung was supported by (4) No. RS-2022-00155911: Artificial Intelligence Convergence Innovation Human Resources Development(Kyung Hee University), (5) No. RS-2024-00509257: Global AI Frontier Lab), and (6) the MSIT(Ministry of Science and ICT), Korea, under the ITRC(Information Technology Research Center) support program(IITP-2024-RS-2024-00438239).

\bibliography{acl}
\clearpage

\appendix

\renewcommand{\thetable}{A\arabic{table}}
\setcounter{table}{0}
\renewcommand{\thefigure}{A\arabic{figure}}
\setcounter{figure}{0}

\section{Appendix}

\subsection{Ablation Study of Top-k documents on MMLU-Med}
\begin{figure}[t]
\centering
\includegraphics[width=\columnwidth]{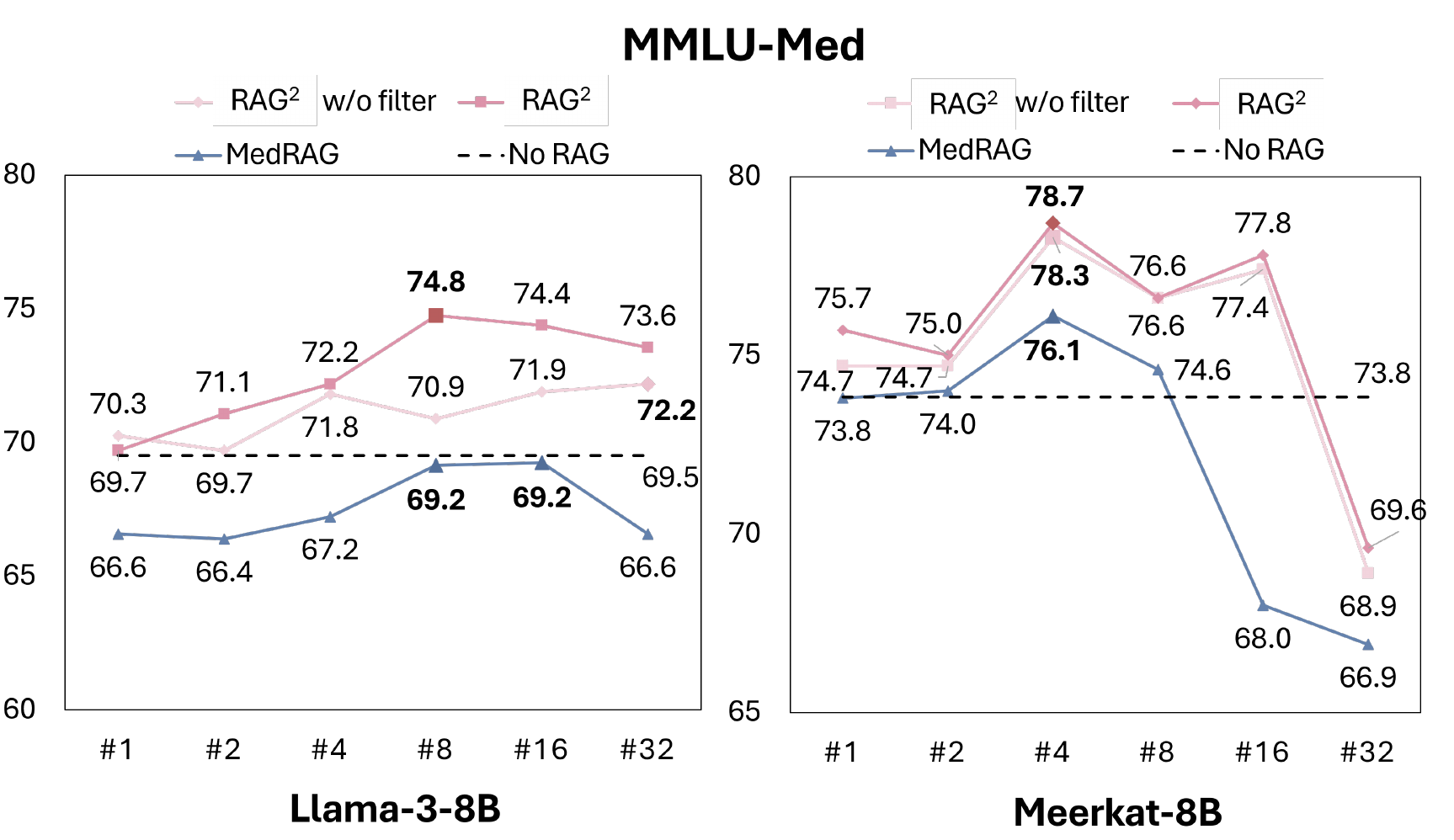}
\caption{
Accuracy (y-axis) of Llama-3-8B-Instruct and Meerkat-7B with/without different RAG models when varying the number of top-k documents (x-axis) on the MMLU-Med dataset. 
Since MMLU-Med only consists of the test set, we used the filtering model trained on MedMCQA during evaluation on MMLU-Med. 
The accuracy results indicate that the effective filter trained from a different dataset (MedMCQA) performs well across other dataset (MMLU-Med).
}
\label{figure:medmmlu}
\end{figure}

Figure~\ref{figure:medmmlu} shows the performance of baseline models with and without RAG methods as the number of retrieved documents is varied.
Note that our filtering model is trained using MedMCQA since the MMLU-Med training set does not exist.
Similar to the results observed in MedQA and MedMCQA (Figure~\ref{figure:topk}), our RAG$^2$ model achieves the best performance, demonstrating the effectiveness of our approach even in out-of-domain distributions.

\subsection{Datasets}
\paragraph{MedQA~\cite{jin2021disease}} MedQA is a multilingual multiple-choice QA dataset derived from professional medical board exams. It encompasses three languages: English, simplified Chinese, and traditional Chinese. The English subset, MedQA-US, specifically comprises USMLE questions.
\paragraph{MedMCQA~\cite{pal2022medmcqa}}  MedMCQA is a comprehensive multiple-choice question-answering dataset designed for medical domain evaluation. It comprises over 194K curated multiple-choice questions derived from AIIMS and NEET PG entrance exams. These questions span across 2.4k healthcare topics and 21 medical subjects, offering a broad coverage of the medical field. 
\paragraph{MMLU-Med~\cite{hendrycksmeasuring}}  MMLU-Med is a specialized subset of the Massive Multitask Language Understanding (MMLU) benchmark, which is first introduced by~\citet{singhal2023large}. 
While the full MMLU benchmark encompasses 57 diverse tasks, MMLU-Med is subset of six biomedical-related domains: anatomy, clinical knowledge, professional medicine, human genetics, college medicine, and college biology.

\begin{table}[t]
    \centering
    \footnotesize
    \begin{tabular}{lrrr}
    \toprule
    \textbf{Corpus} & \textbf{\# Docs} & \textbf{\# documents} & \textbf{Index Size} \\ 
    \midrule
    \multicolumn{4}{l}{\textbf{MedCorp}~\cite{xiong-etal-2024-benchmarking}} \\ 
    \midrule
    PubMed& 23.9M & 23.9M & 211GB \\ 
    Wikipedia & 6.5M & 29.9M & 262GB \\
    StatPearls & 9.3k & 301.2k & 3.0GB \\ 
    Textbooks & 18 & 125.8k & 1.1GB \\
    \midrule
    Total & 30.4M & 54.2M & 477.2GB \\
    \midrule
    \multicolumn{4}{l}{\textbf{Ours}} \\ 
    \midrule
    PubMed& 36.5M & 69.7M & 400GB \\ 
    PMC & 1.1M  & 46.3M & 160GB \\ 
    CPG & 35.7k  & 607.0k   & 3.5GB  \\ 
    Textbooks & 18 & 134.0k & 0.7GB \\
    \midrule
    Total & 37.6M & 116.7M & 564.2GB \\
    \bottomrule
    \end{tabular}
    \caption{
    The statistics of MedCorp~\cite{xiong-etal-2024-benchmarking} and our retrieval corpus.
    CPG represents the clinical guidelines collected by~\citet{chen2023meditron}.
    Note that the number of documents can vary even for the same corpus, as it depends on how documents are divided into documents.
    Additionally, the indexing for MedCorp was done using four different retrievers, while we only used a single retriever, MedCPT, which resulted in a difference in index size.
    }
\label{tab:corp_stats}
\end{table}

\subsection{Implementation Details}

\paragraph{Retrieval}
For our information retrieval tasks, we used the MedCPT retriever and the reranker. The corpus employed in this study is the same as that used in Self-BioRAG~\cite{jeong2024improving}, an earlier work that preceded MedRAG. This corpus includes PubMed abstracts, PMC full texts, and clinician practical guidelines~\citet{chen2023meditron}. (used for training Meditron, with 8 out of 16 publicly available), and 18 medical textbooks.
Unlike MedRAG, which utilizes a different corpus called MedCorp—comprising PubMed, textbooks, StatPearls,\footnote{\url{https://www.statpearls.com/}} and Wikipedia—we chose to retain the Self-BioRAG corpus due to its already proven effectiveness. Although PubMed provides only abstracts, which lack detailed discussions or results, we supplemented this with PMC full texts to provide a more comprehensive dataset.
To ensure comprehensive coverage and prevent the inadvertent truncation of context, we applied a sliding window mechanism with overlap for chunking the documents. 
    
\paragraph{Training}
We trained the filtering model on a single NVIDIA H100 GPU with 80GB memory over 40 epochs, using a learning rate of 3e-5 and a per-device batch size of 16. We selected a few candidate models from the validation set, as performance converged after certain epochs, and evaluated them on the test set. 

\paragraph{Inference}
For inference, the same GPU was used with vLLM~\cite{kwon2023efficient} for faster processing with LLMs. Greedy decoding was employed with a temperature set to 0 to minimize randomness, but some variability persisted due to hardware/software factors. Notably, GPT-4o still exhibited some randomness under these conditions.

\subsection{Open-ended Clinical Questions}

We test our method on multiple-choice QA datasets, including MedQA, MedMCQA, and MMLU-Med.
While these benchmarks are valuable for assessing the medical knowledge of LLMs, they do not fully capture the complexity of real-world clinical scenarios. To address this, we further evaluated our method in a setting that more closely resembles actual clinical environments.
We used the ClinicalQA25 dataset~\cite{zakka2024almanac}, which consists of 25 queries from clinicians.


\subsubsection{Metrics}
We used the following two metrics to measure the models' long-form responses: ROUGE-L and BERTScore.

\paragraph{ROUGE-L~\cite{rouge2004package}} 
ROUGE-L is based on the longest common subsequence (LCS) between the candidate text \( C \) and the reference text \( R \), and is computed as follows:
\begin{equation*}
\begin{aligned}
& \text{ROUGE-L Precision}(C, R) = \frac{LCS(C, R)}{|C|},
\\
& \text{ROUGE-L Recall}(C, R) = \frac{LCS(C, R)}{|R|},\\
& \text{ROUGE-L F1}(C, R) = 2 \times \frac{\text{Precision} \times \text{Recall}}{\text{Precision} + \text{Recall}}.
\end{aligned}    
\end{equation*}

\begin{figure}[t]
\centering
\includegraphics[width=\columnwidth]{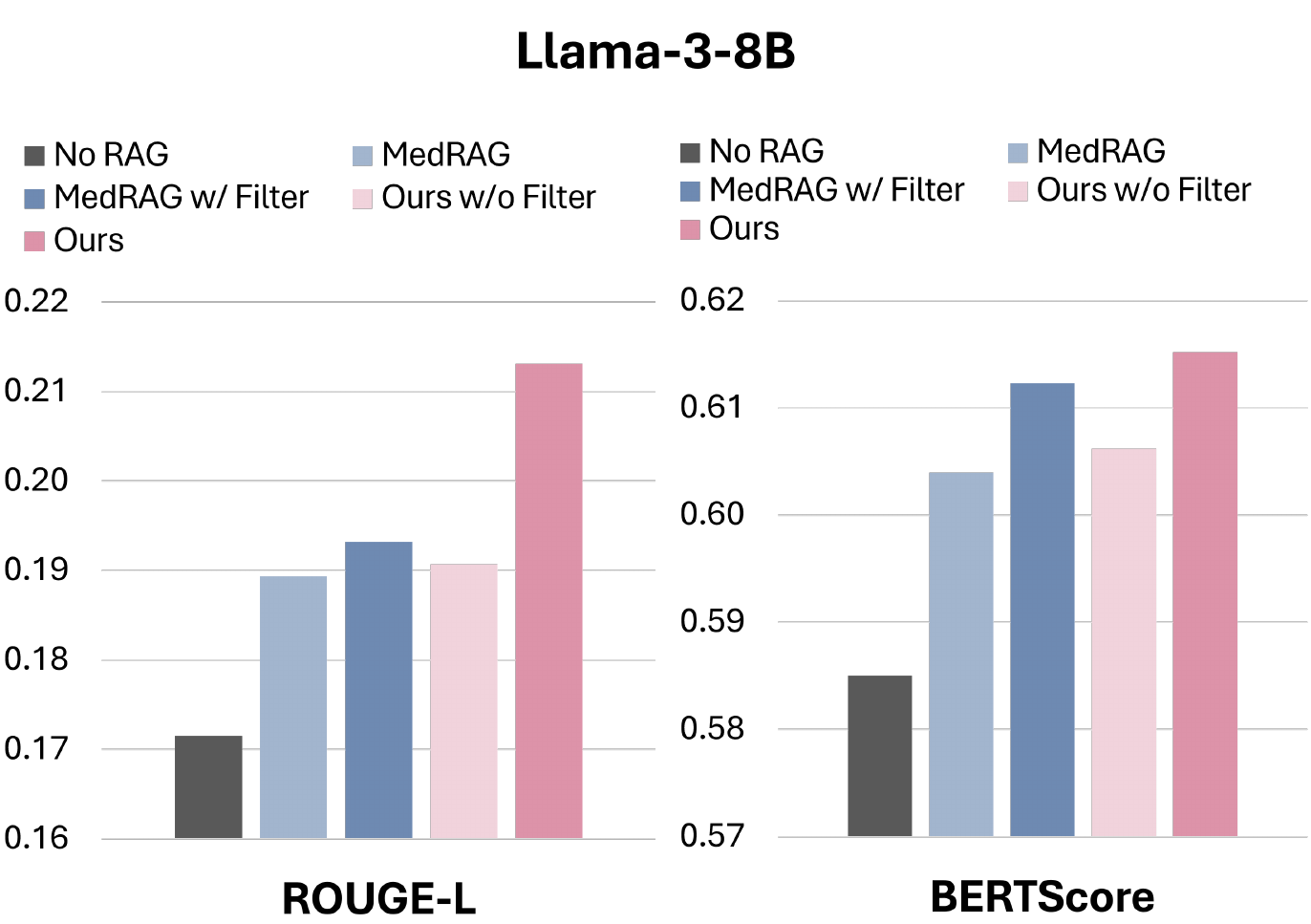} 
\caption{
Performance of Llama-3-8B-Instruct using different RAG pipelines on ClinicalQA25, which represents a real-world clinical scenario. 
}
\label{figure:longform}
\end{figure}

\paragraph{BERTScore~\cite{zhangbertscore}} 
BERTScore computes precision, recall, and F1 score based on cosine similarity between token embeddings, defined as follows:
\begin{equation*}
\resizebox{\columnwidth}{!}{$
\begin{aligned}
& \text{Precision}(C, R) = \frac{1}{|C|} \sum_{c \in C} \max_{r \in R} \text{cosine\_similarity}(c, r),\\
& \text{Recall}(C, R) = \frac{1}{|R|} \sum_{r \in R} \max_{c \in C} \text{cosine\_similarity}(r, c),\\
& \text{BERTScore}(C, R) = 2 \times \frac{\text{Precision} \times \text{Recall}}{\text{Precision} + \text{Recall}}.
\end{aligned}
$}
\end{equation*}

BERTScore is based on the similarity of word embeddings, allowing it to capture semantic meaning more effectively than ROUGE-L.


\subsubsection{Results}
Figure~\ref{figure:longform} demonstrates that RAG, including both our proposed approach and MedRAG, outperforms the model-alone baseline. This suggests that RAG is effective in retrieving accurate and relevant information for open-ended questions. Additionally, the integration of our filtering mechanism seems to further improve performance, potentially enhancing retrieval accuracy.
This experiment demonstrates that our filtering model, although trained on multiple-choice QA, still shows effectiveness in long-form and real-world clinical queries. In future work, we plan to further validate the broader applicability of our approach.

\begin{figure*}[t]
\centering
\includegraphics[width=0.95\textwidth]{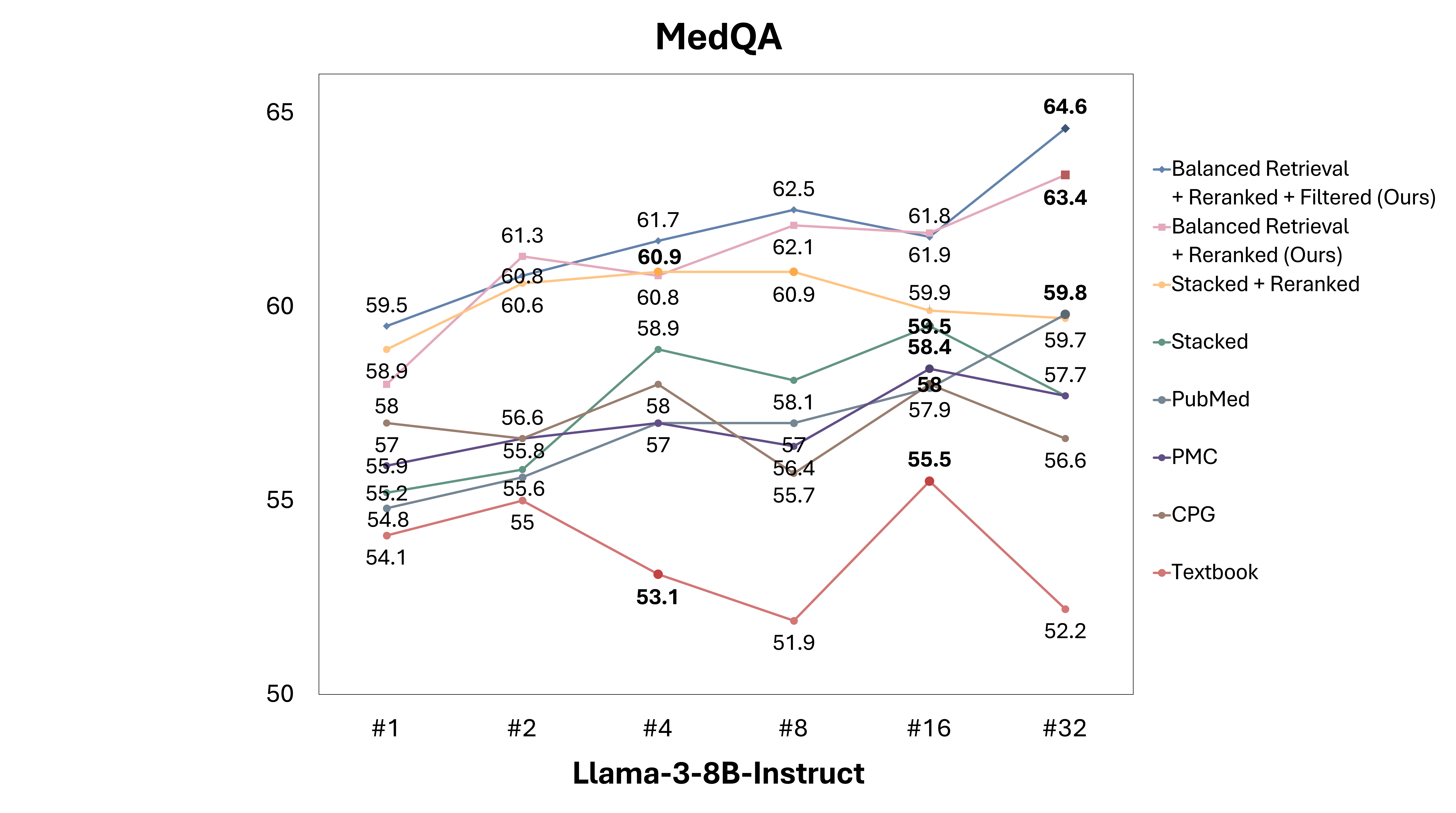} %
\caption{
Comparison of retrieval strategies: independent corpus retrieval, stacked retrieval (following MedRAG~\cite{xiong-etal-2024-benchmarking}), reranked stacked retrieval, and our proposed balanced retrieval, with and without rationale-guided filtered retrieval. The balanced approach consistently outperforms others, highlighting the importance of considering corpus-specific properties in medical information retrieval. The highest scores for each retrieval strategy are marked in bold.
}
\label{figure:corpus_ablation}
\end{figure*}

\subsection{Balanced Retrieval}

To validate the effectiveness of balanced retrieval in medical QA, we conduct corpus ablation studies and compare three retrieval strategies: independent corpus retrieval, stacked retrieval following MedRAG~\cite{xiong-etal-2024-benchmarking} with and without reranking, and our balanced method, with and without filtering (Figure~\ref{figure:corpus_ablation}). 

Our experiments reveal several important findings:
(1) Independent corpus retrieval consistently underperformed, confirming that no single corpus contained sufficient information to cover diverse medical questions.
(2) While the stacked retrieval broadens coverage by combining all sources, it shows lower performance than our balanced retrieval method. This suggests that simply combining corpora can lead to suboptimal results. Even after reranking, the stacked approach does not outperform our balanced retrieval method.
(3) Our balanced retrieval, especially when combined with filtering, consistently achieves the best performance across all retrieval scenarios. This demonstrates that deliberately maintaining representation from diverse sources while filtering out unhelpful documents is crucial for effective medical information retrieval.

These results strongly support the importance of a balanced approach to multi-corpus retrieval in medical QA. By explicitly accounting for the unique properties and contributions of each corpus, our method achieves more reliable and comprehensive information retrieval than approaches that rely on simple corpus combination or single-source retrieval.


%







\end{document}